\documentclass{article}
\usepackage[utf8]{inputenc}
\usepackage{graphicx}
\usepackage{caption}
\usepackage{subcaption}

\graphicspath{ {imgs/} }

\title{Bootstrapping Face Detection with Hard Negative Examples}
\author{Shaohua Wan \quad Zhijun Chen \quad Tao Zhang \quad Bo Zhang \quad Kong-kat Wong \ \\ 
        Xiaomi Inc.\\
        \{wanshaohua, chenzhijun, tao.zhang, zhangbo, kkwong\}@xiaomi.com }

\begin{document}

\maketitle

\begin{abstract}
    Recently significant performance improvement in face
    detection was made possible by deeply trained convolutional
    networks. In this report, a novel approach for training
    state-of-the-art face detector is described. The key is
    to exploit the idea of hard negative mining and iteratively
    update the Faster R-CNN based face detector with the hard
    negatives harvested from a large set of background examples.
    We demonstrate that our face detector outperforms state-of-the-art 
    detectors on
    the FDDB dataset, which is the de facto standard for 
    evaluating face detection algorithms.
\end{abstract}

\section{Introduction}
\label{sec:intro}

Face detection is one of the most widely researched topics in computer
vision, and has found successful applications in face verification
\cite{deepid10000,Taigman:2014:DCG:2679600.2680208}, face
tracking \cite{Seong:2006:EDK:2173493.2173541,34394}, face recognition
\cite{journals/corr/abs-1108-1353,parkhi15deep}, etc. However, it remains a challenging
problem due to the appearance variations caused by changes in viewpoints,
occlusion, facial expression, illumination and cosmetics, etc.

Since the remarkable success of the deep Convolutional Neural Network
(CNN) \cite{NIPS2012_4824} in image classification on the ImageNet Large Scale Visual
Recognition Challenge (ILSVRC) 2012, numerous efforts have been made to
port CNN to face detection.
Faster R-CNN \cite{NIPS2015_5638}, originally developed as a generic
object detector, was recently demonstrated to produce 
state-of-the-art performance on face detection tasks
\cite{DBLP:journals/corr/JiangL16a}.

Faster R-CNN is trained through a reduction
that converts object detection into an image classification problem. That
is, Faster R-CNN learns to detect objects by scoring a large set of regions
of interest (RoIs) that are sampled from the images. This 
inevitably introduces a challenge that is not seen in standard
image classification problems: the 
set of RoIs is distinguished by a highly imbalanced distribution; the ratio 
between the number of background (bg) RoIs 
(regions not belonging to any object of interest) and the 
number of foreground (fg) RoIs could reach as high as 100:1. 
Given such a high bg/fg ratio, it is virtually impossible to consider
all the region proposals simultaneously. 
Faster R-CNN counteracts data imbalance by
randomly subsampling the background examples and maintaining
a bg/fg ratio of 3:1 in each mini-batch. 
This naive technique mitigates data imbalance to some extent,
but more advanced strategy is required to further improve object 
detection performance.

Data imbalance has long been a problem for region-based object
detectors. It is common that training data contains
an overwhelming number of negative examples, most of which
are easy ones for the detector and only a few are difficult.
A standard solution, known as hard negative mining, is to iteratively grow, or bootstrap, a small set of 
negative examples by selecting those negatives for 
which the detector triggers a false positive alarm 
\cite{kksung_tpoggio,Dalal05histogramsof,Felzenszwalb:2010:ODD:1850486.1850574, He2014}. 
This strategy leads to an iterative training 
algorithm that alternates between learning the detection model given the 
current set of examples, and then using the learned model to find
hard negatives to add to the training set. 

In this report, we adopt this simple yet effective hard negative mining 
technique for training state-of-the-art face detection models
based on Faster R-CNN. 
We demonstrate that it yields significant
boosts in detection performance on face detection benchmarks 
like FDDB \cite{fddbTech}.

\section{Related Work}

Hard negative mining was first introduced by Sung and Poggio
\cite{kksung_tpoggio} to select high quality examples for function
approximation learning tasks. Since then, hard negative mining
has been widely used to train region-based object detectors
\cite{Dalal05histogramsof,Felzenszwalb:2010:ODD:1850486.1850574, DBLP:conf/cvpr/GirshickDDM14}.
Recently significant performance boosts in object detection were
made possible by CNN. CNN-based object detectors like
R-CNN \cite{DBLP:conf/cvpr/GirshickDDM14} and SPPnet
\cite{He2014} employ SVMs trained with hard negative mining to
detect objects. 
\cite{DBLP:journals/corr/ShrivastavaGG16} propose an online
hard example mining algorithm for Fast R-CNN, which uses
a data sampling strategy that favors those
with high training loss. 
\cite{he16eccv_frcnn_hnm} adopt cascaded Boosted Forest, which perform effective 
hard negative mining and sample re-weighting,
to classify the region proposals generated by RPN.
Our work is unique from previous works in that
we harvest hard negative examples from the output of
Fast R-CNN, which are used to jointly re-train both RPN and 
Fast R-CNN.

\section{Bootstrapping Faster R-CNN with Hard Negative Examples}
\subsection{Overview of Faster R-CNN}

Faster R-CNN consists of two modules. The first, called the Region
Proposal Network (RPN), is a fully convolutional network for generating
regions of interest (RoIs) that denote the possible presence of objects.
The second module is the Fast R-CNN, whose purpose is to classify the
RoIs and refine their position and scales. To save computing resources,
RPN and Fast R-CNN share the same convolution layers up to their own fully
connected layers. The mini-batch in each SGD step consists of foreground RoIs and
background RoIs sampled from each image.

\textbf{Foreground RoIs} Foreground RoIs are those whose intersection over 
union (IoU) overlap with a ground-truth bounding box is greater
than \textsf{th}$_{\textsf{fg}}$. \textsf{th}$_{\textsf{fg}}=0.5$ 
is a fairly standard design choice, which finds widespread
use in R-CNN, SPPnet, and Fast R-CNN. 
The same setting is used in Faster R-CNN.

\textbf{Background RoIs} Background RoIs are those whose maximum IoU with 
ground truth is in the interval [\textsf{th}$_{\textsf{bg}}$, 0.5). 
\textsf{th}$_{\textsf{bg}}=0.1$ is used by both Fast R-CNN and SPPnet.
The idea is that regions 
with some overlap with the ground truth are more likely to be
hard negatives \cite{girshick15fastrcnn}. 

\textbf{Balancing bg-fg RoIs} To handle the data imbalance,
Faster R-CNN \cite{girshick15fastrcnn, NIPS2015_5638} 
fix the bg/fg ratio 
in each mini-batch to a target of 3:1 by undersampling the 
bg RoIs at random. As mentioned in Section~\ref{sec:intro},
detectors trained in this manner are susceptible to hard negative
examples and tend to produce more false alarms.

\subsection{Hard Negative Mining}

The key idea of hard negative mining is to construct an initial training 
set consisting of positive RoIs and random subset of negative RoIs.
The face detection model is learned with this training set and subsequently
applied to all negatives to harvest false positives.
The false positives are then added to the training set and
then the model is trained again. This process is iterated
several times until satisfaction.

We label a detected region as a hard negative
if its maximum IoU with any groundtruth face annotation is
less than 0.5.
When the model is re-trained with the hard negatives added,
we still maintain a bg/fg ratio of 3:1 in each mini-batch
while ensuring that the hard negatives harvested by 
the previously trained model are selected. 

One should note that two optimization schemes exist for solving 
a Faster R-CNN network: (1) a 4-step alternating optimizaiton algorithm 
that alternates between fine-tuning for RPN and then fine-tuning
Fast R-CNN; (2) an approximate joint optimization algorithm that
solves Faster R-CNN as a single network.
We take the latter optimization scheme for its simplicity and speed.

\begin{figure}
    \centering
    \begin{subfigure}[b]{0.95\textwidth}
        \includegraphics[width=\textwidth]{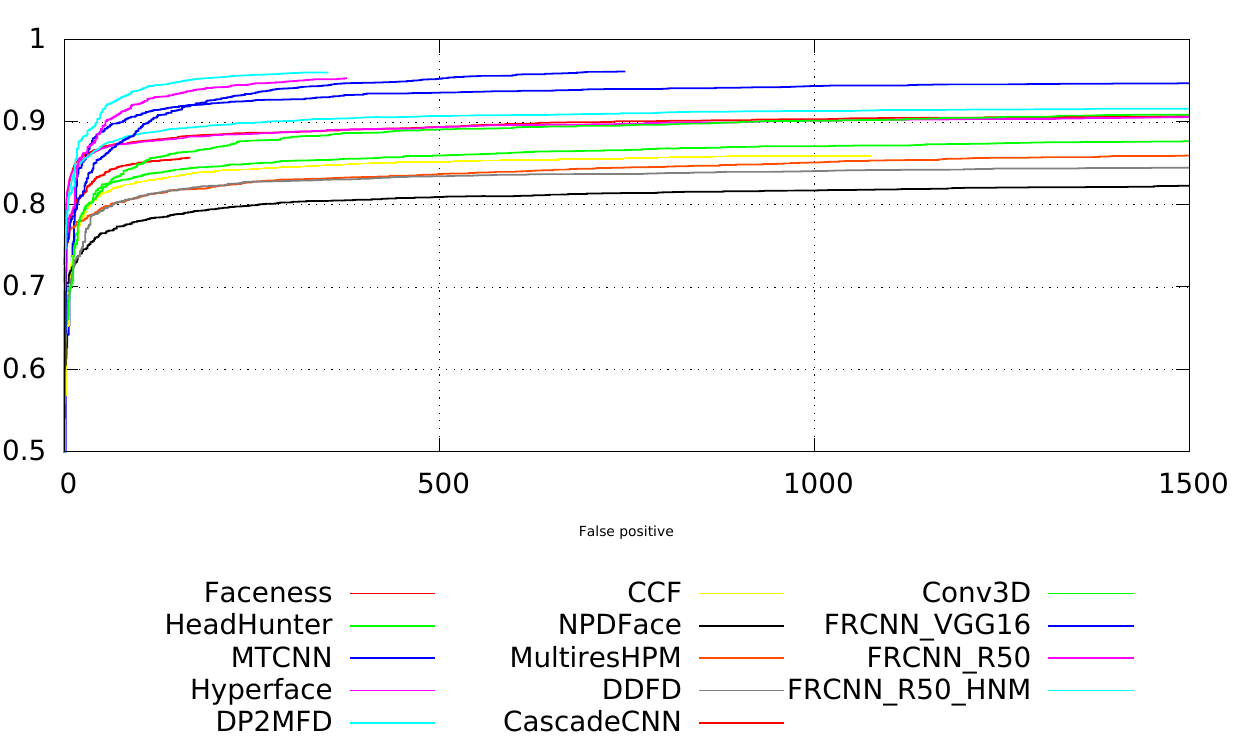}
        \caption{ROC curves with discrete scores.}
        \label{fig:fddb-disc}
    \end{subfigure}
    \begin{subfigure}[b]{0.95\textwidth}
        \includegraphics[width=\textwidth]{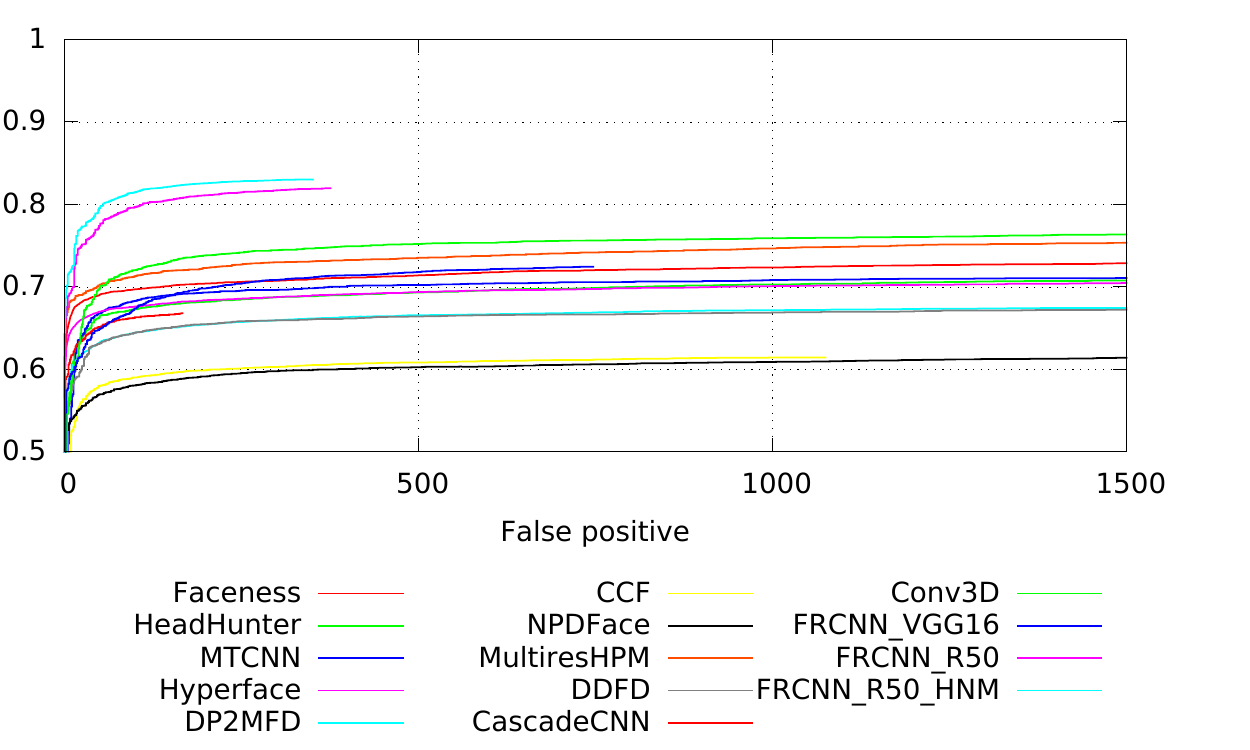}
        \caption{ROC curves with continuous scores.}
        \label{fig:fddb-cont}
    \end{subfigure}
    \caption{The face detection performance of various methods on the
    FDDB dataset.
    FRCNN\_VGG16 is the Faster R-CNN face detector
with the VGG16 network architecture, as described in
\cite{DBLP:journals/corr/JiangL16a}.
FRCNN\_R50\_HNM is the Faster R-CNN face detector with the
ResNet-50 network architecture, trained with hard negative mining.
We also obtain results for FRCNN\_R50, which is the 
Faster R-CNN face detector with the ResNet-50 network architecture,
trained without hard negative mining.}
    \label{fig:fddb}
\end{figure}

%
%

\section{Experiments}
In this section, we compare various detection algorithms
and report their performance on the FDDB \cite{fddbTech} face detection dataset.

\subsection{FDDB}

\begin{itemize}
  \item FDDB has a
  total of 5,171 faces in 2,845 images,
  with a wide range of detection difficulties including occlusions, 
  difficult poses, and low resolution and out-of-focus faces.
  As specified in \cite{fddbTech}, we conduct
  10-fold cross-validation experiments, and use
  the FDDB evaluation software to generate two performance
  curves: (a) discrete ROC curve, and (b) continuous ROC curve. To
  generate the discrete ROC curve, 
  each detection is assigned a binary match/non-match label. The
  continuous ROC curve associates a real-valued score with each
  detection based on the overlap between the detected and the
  annotated regions.

\end{itemize}

We train and test Faster R-CNN on single-scale images.
The images are rescaled such that the shorter side is $s = 600$ pixels.
ResNet-50 \cite{he15deepresidual} is used as the network architecture
and is initialized by the pre-trained
ImageNet classification model.
Approximate joint optimization mode is used to train Faster R-CNN,
with a base learning rate of 0.001 for the first 50k iterations
and a reduced learning rate of 0.0001 for 20k more iterations.

Our hard negative mining proceeds in two rounds.
In the first round, we train the Faster R-CNN face detection model just
as normal. In the second round, 
hard negatives are harvested from the training images using 
face detection model obtained from the first round. The face
detection model is re-trained with the hard negatives added
to each mini-batch.

We compare the proposed face detection algorithm with several
other state-of-the-art algorithms listed on the official FDDB
website, and plot the discrete and continuous ROC curves
in Fig.~\ref{fig:fddb}. FRCNN\_VGG16 is the Faster R-CNN face detector
with the VGG16 network architecture, as described in \cite{DBLP:journals/corr/JiangL16a}.
FRCNN\_R50\_HNM is the Faster R-CNN face detector with the
ResNet-50 network architecture, trained with hard negative mining.
We also obtain results for FRCNN\_R50, which is the 
Faster R-CNN face detector with the ResNet-50 network architecture,
trained without hard negative mining. As can be seen,
for both discrete and continuous ROC curves, FRCNN\_VGG16,
FRCNN\_R50, and 
FRCNN\_R50\_HNM are the three best performing methods for
face detection on FDDB, demonstrating the effectiveness of Faster R-CNN
for face detection tasks. FRCNN\_R50 outperforms FRCNN\_VGG16
mainly due to the superiority of ResNet-50 over VGG16 in learning a deeper, better
image feature representation. 
Moreover, hard negative mining is able to boost the
performance of FRCNN\_R50 by a large margin, as indicated
by the ROC curve of FRCNN\_R50\_HNM.
For detailed performance statistics, we refer readers to 
the official FDDB website
\footnote{http://vis-www.cs.umass.edu/fddb/results.html}.


\section{Conclusion}
In this report, we present the latest face detection results on the
FDDB dataset, demonstrating that significant performance gains
over state-of-the-arts
can be achieved by learning Faster R-CNN face detection models with
hard negative mining. 

It is worth noting that, our face detection model, though being
robust to difficult imaging conditions such as occlusions, 
difficult poses, and low-resolution faces, is less resilient
to faces of small sizes. This is mainly due to the low-resolution
feature maps of the last convolution layer on small faces.
One possible solution is to pool feature maps of shallow
but high-resolution convolution
layers, and this is subject to future investigation.


\begin{thebibliography}{10}

\bibitem{Dalal05histogramsof}
Navneet Dalal and Bill Triggs.
\newblock Histograms of oriented gradients for human detection.
\newblock In {\em IEEE Conference on Computer Vision and Pattern Recognition},
  pages 886--893, 2005.

\bibitem{Felzenszwalb:2010:ODD:1850486.1850574}
Pedro~F. Felzenszwalb, Ross~B. Girshick, David McAllester, and Deva Ramanan.
\newblock Object detection with discriminatively trained part-based models.
\newblock {\em IEEE Trans. Pattern Anal. Mach. Intell.}, 32(9):1627--1645,
  September 2010.

\bibitem{girshick15fastrcnn}
Ross Girshick.
\newblock Fast {R-CNN}.
\newblock In {\em International Conference on Computer Vision}, 2015.

\bibitem{DBLP:conf/cvpr/GirshickDDM14}
Ross~B. Girshick, Jeff Donahue, Trevor Darrell, and Jitendra Malik.
\newblock Rich feature hierarchies for accurate object detection and semantic
  segmentation.
\newblock In {\em IEEE Conference on Computer Vision and Pattern Recognition},
  pages 580--587, 2014.

\bibitem{He2014}
Kaiming He, Xiangyu Zhang, Shaoqing Ren, and Jian Sun.
\newblock {\em Spatial Pyramid Pooling in Deep Convolutional Networks for
  Visual Recognition}, pages 346--361.
\newblock 2014.

\bibitem{he15deepresidual}
Kaiming He, Xiangyu Zhang, Shaoqing Ren, and Jian Sun.
\newblock Deep residual learning for image recognition.
\newblock In {\em arXiv prepring arXiv:1506.01497}, 2015.

\bibitem{fddbTech}
Vidit Jain and Erik Learned-Miller.
\newblock Fddb: A benchmark for face detection in unconstrained settings.
\newblock Technical Report UM-CS-2010-009, University of Massachusetts,
  Amherst, 2010.

\bibitem{DBLP:journals/corr/JiangL16a}
Huaizu Jiang and Erik~G. Learned{-}Miller.
\newblock Face detection with the faster {R-CNN}.
\newblock {\em CoRR}, abs/1606.03473, 2016.

\bibitem{34394}
Minyoung Kim, Sanjiv Kumar, Vladimir Pavlovic, and Henry~A. Rowley.
\newblock Face tracking and recognition with visual constraints in real-world
  videos.
\newblock In {\em IEEE Conference on Computer Vision and Pattern Recognition},
  2008.

\bibitem{NIPS2012_4824}
Alex Krizhevsky, Ilya Sutskever, and Geoffrey~E. Hinton.
\newblock Imagenet classification with deep convolutional neural networks.
\newblock In {\em Advances in Neural Information Processing Systems}, pages
  1097--1105. 2012.

\bibitem{journals/corr/abs-1108-1353}
K.~Susheel Kumar, Vijay~Bhaskar Semwal, and R.~C. Tripathi.
\newblock Real time face recognition using adaboost improved fast pca
  algorithm.
\newblock {\em CoRR}, abs/1108.1353, 2011.

\bibitem{parkhi15deep}
O.~M. Parkhi, A.~Vedaldi, and A.~Zisserman.
\newblock Deep face recognition.
\newblock In {\em Proceedings of the British Machine Vision Conference}, 2015.

\bibitem{NIPS2015_5638}
Shaoqing Ren, Kaiming He, Ross Girshick, and Jian Sun.
\newblock Faster r-cnn: Towards real-time object detection with region proposal
  networks.
\newblock In C.~Cortes, N.~D. Lawrence, D.~D. Lee, M.~Sugiyama, and R.~Garnett,
  editors, {\em Advances in Neural Information Processing Systems}, pages
  91--99. 2015.

\bibitem{Seong:2006:EDK:2173493.2173541}
Chi-Young Seong, Byung-Du Kang, Jong-Ho Kim, and Sang-Kyun Kim.
\newblock Effective detector and kalman filter based robust face tracking
  system.
\newblock In {\em Proceedings of the First Pacific Rim Conference on Advances
  in Image and Video Technology}, pages 453--462, 2006.

\bibitem{DBLP:journals/corr/ShrivastavaGG16}
Abhinav Shrivastava, Abhinav Gupta, and Ross~B. Girshick.
\newblock Training region-based object detectors with online hard example
  mining.
\newblock {\em CoRR}, abs/1604.03540, 2016.

\bibitem{deepid10000}
Yi~Sun, Xiaogang Wang, and Xiaoou Tang.
\newblock Deep learning face representation from predicting 10,000 classes.
\newblock In {\em IEEE Conference on Computer Vision and Pattern Recognition},
  June 2014.

\bibitem{kksung_tpoggio}
K.-K. Sung and T.~Poggio.
\newblock Learning and example se-lection for object and pattern detection.
\newblock {\em MIT A.I. Memo}, 1521, 1994.

\bibitem{Taigman:2014:DCG:2679600.2680208}
Yaniv Taigman, Ming Yang, Marc'Aurelio Ranzato, and Lior Wolf.
\newblock Deepface: Closing the gap to human-level performance in face
  verification.
\newblock In {\em IEEE Conference on Computer Vision and Pattern Recognition},
  pages 1701--1708, 2014.

\bibitem{he16eccv_frcnn_hnm}
Liliang Zhang, Liang Lin, Xiaodan Liang, and Kaiming He.
\newblock Is faster r-cnn doing well for pedestrian detection?
\newblock In {\em arXiv prepring arXiv:1607.07032}, 2016.

\end{thebibliography}

\bibliographystyle{plain}

\end{document}